\documentclass[letterpaper, 10 pt, conference]{ieeeconf}  
\usepackage[T1]{fontenc}
\usepackage{amsmath}
\usepackage{amssymb}
\usepackage{graphicx}
\usepackage{booktabs}
\usepackage{multirow}
\usepackage{float}

\IEEEoverridecommandlockouts                              

\title{\LARGE \bf
RM-WAM: Reliable Memory-Augmented World Action Models\\for Long-Horizon Robot Manipulation
}

\author{
\parbox{\linewidth}{
\centering
Cui Miao$^{1,2}$,
Junhao Yu$^{3}$,
Ping Liu$^{2}$,
Ning Yang$^{4}$,
Xuhua Chen$^{2}$,\\
He Zheng$^{2}$,
Xiaodong Wang$^{1}$,
Tao Zhang$^{2}$$^{*}$\\[3pt]
\small
$^{1}$National University of Defense Technology \quad
$^{2}$Magic-Lab Team, Magiclab Robotics Inc.\\
$^{3}$Beijing Institute of Technology \quad
$^{4}$Nanjing University
}
}

\begin{document}

\maketitle
\thispagestyle{empty}
\pagestyle{empty}

\begin{abstract}

World Action Models (WAMs) improve robotic manipulation by jointly predicting future visual states and actions, yet most existing methods make decisions primarily from the current observation. This limits their reliability in long-horizon tasks, where visually similar states may correspond to different interaction histories and task stages. Moreover, directly incorporating historical observations can introduce stale, stage-mismatched, or error-contaminated information, leading to inaccurate world prediction and accumulated closed-loop errors. We propose RM-WAM, a reliable memory-augmented World Action Model that treats memory as a context-dependent predictive resource rather than universally useful historical context. RM-WAM represents history using complementary short-term visual state memory and latent event memory, which respectively capture recent interaction dynamics and longer-term state transitions. A counterfactual memory learning scheme varies the accessible history while keeping the current inputs and prediction targets fixed, enabling the model to learn a context-dependent utility estimate for historical memory. Based on this estimate, a utility-aware gating mechanism adaptively controls memory injection into the world–action backbone. Experiments demonstrate that RM-WAM highlights the importance of reliable and selective memory utilization in long-horizon tasks.
\end{abstract}

\section{INTRODUCTION}

By further training pretrained vision-language models on embodied data, Vision-Language-Action models (VLAs) are becoming an important foundation model paradigm for robotics and show strong abilities in task understanding and generalization~\cite{kim2024openvla,team2024octo,intelligence2025pi_,li2024cogact}. However, existing VLAs focus on  predicting the next action from the current observation directly. This local decision-making paradigm has clear limitations in long-horizon tasks, because each action not only determines the immediate behavior but also affects subsequent environment states and task progress. World Action Models (WAMs) are designed to bridge this gap~\cite{kim2026cosmos,ye2026gigaworld,ye2026world,bi2026motus,yuan2026fast}. By modeling action-conditioned state transitions and future observations, WAMs predict both what action the robot should take and how the world changes after the action is executed. However, reliable prediction of world evolution depends not only on the current observation, but also on a continuous understanding of past interactions. This makes historical context particularly important for WAMs in complex long-horizon manipulation, where visually similar observations may correspond to different interaction states and require different future evolutions.

\begin{figure}[t]
    \centering
    \includegraphics[width=0.45\textwidth]{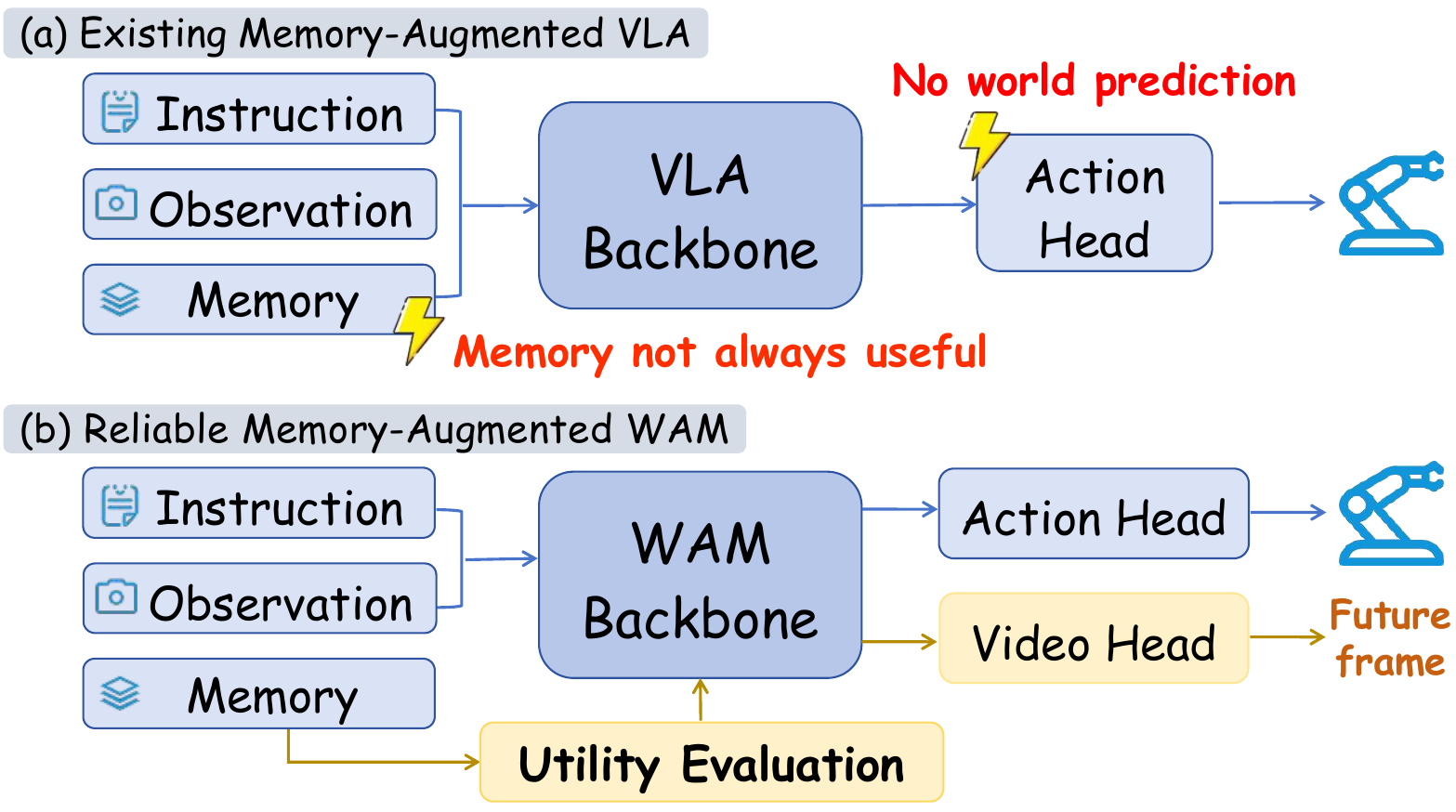}
    \caption{
    \textbf{Motivation of RM-WAM.} (a) Existing memory-augmented VLAs use memory for action prediction, but retrieved memory is not always useful and they do not model future world prediction. (b) RM-WAM evaluates memory utility and incorporates reliable memory into a WAM for both world prediction and action generation.}
    \label{fig:motivation}
\end{figure}

Recent studies use memory mechanisms to improve policy performance in long-horizon manipulation tasks. For example, MemoryVLA~\cite{shi2025memoryvla}, MEM~\cite{torne2026mem}, and other memory-augmented methods~\cite{li2026global,li2026remem,wang2026mem} build perceptual, semantic, or multi-scale memories to provide additional history for long-horizon action generation. However, directly extending these memory mechanisms to WAMs presents two fundamental limitations, as illustrated in Fig.  \ref{fig:motivation}. First, existing memory-augmented VLAs are primarily designed to support action prediction, whereas WAMs additionally model action-conditioned world evolution through future visual-state prediction. Memory in a WAM therefore needs to support not only action generation but also world-state modeling, where inappropriate historical information can directly distort the predicted future. Second, retrieving task-relevant history does not necessarily mean that the retrieved memory is useful for the current world-action prediction. Existing memory mechanisms typically retrieve historical information according to relevance or similarity and then use the retrieved memory as additional context. However, even highly relevant memory may be stale, inconsistent with the current task stage, or contaminated by previous rollout errors. Therefore, WAMs require an additional mechanism to explicitly assess the utility of retrieved memory before incorporating it into the prediction process.

This view is also consistent with cognitive science, where memory is not treated as a passive replay buffer but as an active and context-dependent source of control~\cite{ikier2024memory}. Humans selectively retrieve and reorganize past experience when it supports the current goal, while suppressing or weakening memories that interfere with ongoing behavior. Similarly, historical information in WAMs should not be assumed to be beneficial simply because it is relevant to the task. Its usefulness depends on the current observation, task stage, and prediction objective. We therefore argue that the key challenge in memory-augmented WAMs is not only \emph{which memory to retrieve}, but also \emph{whether the retrieved memory should be used}. This motivates explicitly modeling memory utility for reliable world-action prediction.

To address this challenge, we propose Reliable Memory-augmented World Action Model (RM-WAM), a utility-aware memory framework for long-horizon robot manipulation. We first model history as short-term visual memory and latent event memory according to their different roles in world-state prediction. Short-term visual memory describes local visual changes, providing interaction states that are difficult to infer from current observation. Latent event memory encodes historical transitions in the shared latent space and captures history that is relevant to future world evolution. We then introduce a counterfactual memory learning scheme for WAMs. By keeping the current observation and task instruction fixed while changing the accessible history, the model learns how different memory conditions affect future prediction and action generation, thereby obtaining a context-dependent utility estimate for historical memory under the current state. Based on the learned  utility, the model adaptively controls memory injection before WAM backbone, so that it preserves history that  helps the current prediction while suppressing stale, stage-misaligned, or error-contaminated memory. In this way, memory is no longer treated as extra  context that is useful by default, but as a selective historical intervention in world evolution.

Our contributions are summarized as follows:

\begin{itemize}
    \item We extend memory augmentation to the World Action Models, and show that the key challenge in memory-augmented WAMs is not only storing or retrieving history, but also determining whether memory is consistent with the current world-evolution prediction and whether it has positive utility for future state estimation and closed-loop decision making. 
    \item We propose Reliable Memory-augmented WAM, a utility-aware memory framework for long-horizon robotic manipulation. The proposed method models history at different scales through short-term visual memory and latent event memory, and combines counterfactual memory learning with utility-aware memory adapters to selectively preserve helpful history while suppressing stale, stage-misaligned, or error-contaminated memory.
    \item We evaluate RM-WAM on both simulation benchmarks including LIBERO~\cite{liu2023libero}, memory-dependent RMBench~\cite{chen2026rmbench} and challenging long-horizon real-world task. Compared with representative non-memory and memory-augmented baselines, RM-WAM consistently achieves strong performance, demonstrating the benefit of reliable memory utilization for World Action Models.
\end{itemize}

\vspace{-2mm}
\section{Related Works}

\subsection{Vision-Language-Action Models} Vision-Language-Action models (VLAs) integrate visual observations, language instructions, and robot actions into end-to-end policies \cite{bu2025univla,team2025gigabrain,jang2025contextvla}. Many recent VLAs build on pretrained vision-language models  (VLMs) and are further trained on large-scale robot demonstrations, improving generalization across tasks, environments, and robot embodiments \cite{marafioti2025smolvlm,bai2025qwen3}. Early works such as RT-1~\cite{brohan2022rt} and RT-2~\cite{zitkovich2023rt} show the scalability of Transformer-based policies. Open X-Embodiment~\cite{o2024open}, Octo~\cite{team2024octo}, and OpenVLA~\cite{kim2024openvla} further push multi-embodiment training and efficient downstream adaptation. The $\pi$ family~\cite{black2024pi_0,intelligence2025pi_} further uses diffusion or flow matching to generate continuous action chunks. However, existing VLA methods mainly learn a direct mapping from current observations and language instructions to action sequences, while overlooking how the environment evolves under robot actions. 

\vspace{-2mm}
\subsection{World Action Models} World models learn environment dynamics by predicting future observations or latent states, providing predictive representations for planning and policy learning \cite{cen2025worldvla,li2025hamster,zhang2026world}. The Dreamer series~\cite{hafner2019dream,hafner2023mastering} optimizes policies in a learned latent world and shows the value of imagined trajectories. UniPi~\cite{du2023learning} and Video Language Planning~\cite{du2024video} use future videos for planning and convert visual predictions into robot actions through inverse dynamics or goal-conditioned policies. Recent world action models jointly model future visual states and actions in a unified architecture~\cite{ye2026world,bi2026motus}. However, existing WAMs mainly rely on the current observation and lack effective modeling of historical state. This limits history-dependent prediction in long-horizon tasks with partial observability and stage-dependent behavior.

\vspace{-2mm}
\subsection{Memory-Augmented Robotic Manipulation} Memory is important for handling long-horizon task dependencies in robotic manipulation. Early approaches mainly use recurrent states or temporal attention to aggregate history~\cite{bulatov2022recurrent}. Recent studies use updatable memory bank to extend the temporal context of robot policies~\cite{lei2025robomemory,torne2026mem}. However, directly applying memory-augmented mechanisms to WAMs does not ensure that the retrieved history remains consistent with the current world prediction. Although recent studies improve policy performance by introducing hierarchical, persistent, or higher-capacity memory mechanisms~\cite{sun2026himem,yang2026memorywam}, they primarily focus on organizing and extending historical context, while the context-dependent utility of retrieved memory for current world-action prediction remains less explicitly modeled. Historical information may be stale, stage-mismatched, or corrupted by closed-loop errors. Our RM-WAM therefore treats visual memory as a predictive resource with context-dependent utility rather than uniformly useful historical context.

\begin{figure*}[t]
    \centering
    \includegraphics[width=1\textwidth]{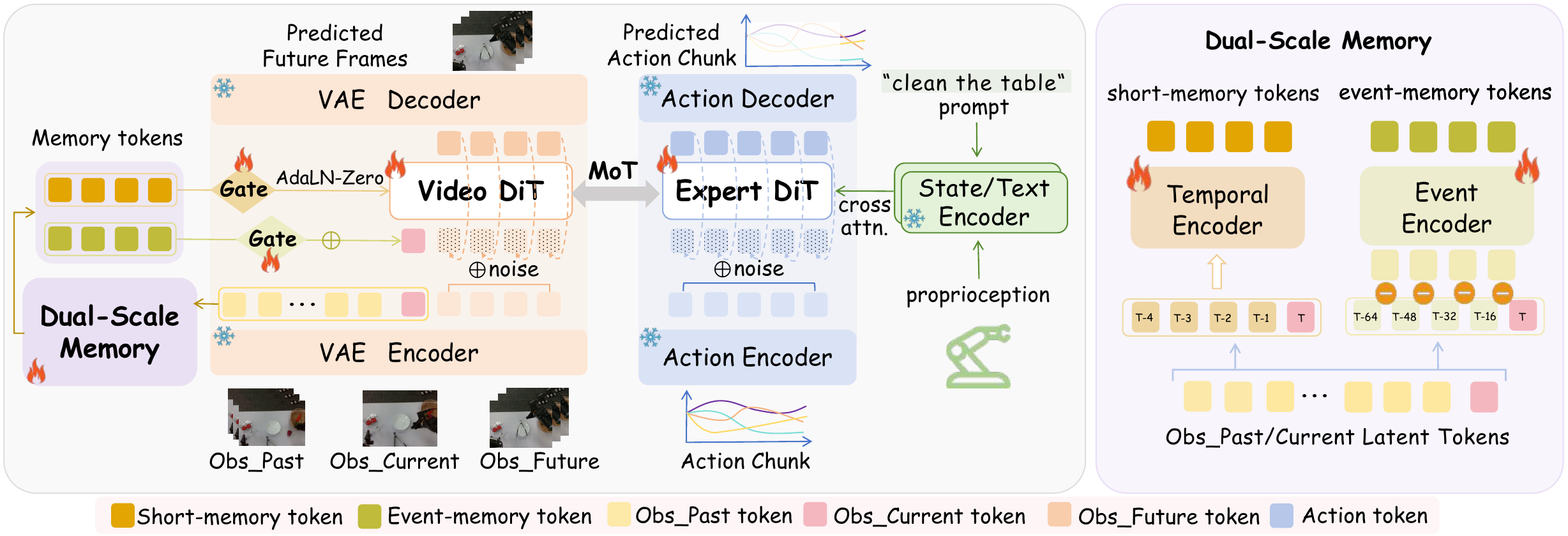}
    \caption{
    \textbf{Overview of RM-WAM.} The left panel shows the overall world-action framework with selective memory injection, while the right panel details the construction of the dual-scale memory.}
    \label{fig:overview}
\end{figure*}

\vspace{-2mm}
\section{Method}

\subsection{Problem Formulation}

A World Action Model (WAM) jointly predicts future visual states and robot actions:
\vspace{-1mm}
\begin{equation}
p_\theta\left(
\mathbf{h}_{t+1:t+H_v},
\mathbf{a}_{t:t+H_a-1}
\mid o_t,c,p_t
\right),
\end{equation}
where $o_t$, $c$, and $p_t$ denote the current visual observation, language instruction, and proprioceptive state, respectively. $\mathbf{a}_{t:t+H_a-1}$ is the future action chunk with horizon $H_a$, $\mathbf{h}_{t+1:t+H_v}$ denotes future visual states, and $H_v$ is the visual prediction horizon. However, the current observation alone is often insufficient for long-horizon tasks. We therefore introduce history memory:
\vspace{-2mm}
\begin{equation}
\mathcal{M}_t
=
\left(
m_{t-\delta_1},\ldots,m_{t-\delta_K}
\right),
\qquad
0<\delta_1<\cdots<\delta_K,
\end{equation}
where $m_{t-\delta_i}$ denotes a memory item from historical timestep $t-\delta_i$, $\delta_i$ is the corresponding temporal offset, and $K$ is the number of memory items. The memory-conditioned prediction is formulated as:
\begin{equation}
p_\theta
(\mathbf{h}_{t+1:t+H_v},
\mathbf{a}_{t:t+H_a-1}
\mid o_t,c,p_t,\mathcal{M}_t).
\end{equation}
We focus on selecting memory according to its utility under the current context, so as to improve future-state and action prediction in long-horizon tasks.



\subsection{Framework Overview}
The overall framework of RM-WAM is illustrated in Fig.~\ref{fig:overview}. Given the current and historical observations, language instruction, and proprioceptive state, RM-WAM first constructs a dual-scale memory from visual latent tokens. A temporal encoder aggregates recent latent states to form short-term memory tokens, while an event encoder represents historical state transitions as event-memory tokens. These memory representations are selectively injected into the video branch through gated adapters, where the gating functions are learned with counterfactual memory supervision to estimate the utility of memory. The video branch adopts Wan2.2-TI2V-5B \cite{wan2025wan} as the backbone, including its VAE encoder-decoder and Video DiT, while the action branch employs an Expert DiT for action prediction. The two branches interact through a Mixture-of-Transformers (MoT) and are conditioned on language and proprioceptive information through the state/text encoder, enabling memory-enhanced joint prediction of future visual states and action chunks.


\subsection{Counterfactual Memory}

To estimate the utility of historical memory under the current context, we construct counterfactual memory conditions by varying only the accessible history while keeping the current observation, task instruction, proprioceptive state and prediction target unchanged. This design isolates the influence of memory from changes in the current inputs and supervision targets. Specifically, $\mathcal{M}_t^{\mathrm{ref}}$ is constructed from the causally aligned history preceding the current timestep within the same episode and serves as the reference memory condition, $\mathcal{M}_t^{\mathrm{stale}}$ is sampled from temporally distant history, representing information that is valid within the episode but less relevant to the current state, $\mathcal{M}_t^{\mathrm{stage}}$ is  constructed by temporally shifting the historical sampling positions within the same episode, producing history that is causally valid but misaligned with the current decision point. This condition simulates temporally mismatched context without requiring explicit stage annotations, $\mathcal{M}_t^{\mathrm{cross}}$ is drawn from another episode of the same task, preserving task-level semantics while breaking the correspondence with the current trajectory. These counterfactual conditions expose the model to historical contexts with different expected levels of predictive reliability. Each condition is associated with a soft utility target \(q_v\in[0,1]\), which serves as a supervision proxy for the relative reliability of memory under the current prediction context, rather than a direct measurement of causal performance gain.

\vspace{-2mm}
\subsection{Dual Memory Representation}

\subsubsection{\textbf{Short Visual Memory}}
Short memory provides recent visual for WAM and supplements local interaction states that are difficult to infer from a single current frame, such as manipulator occlusion, small object displacement, and short-term motion trends. Directly concatenating all historical tokens increases sequence length and introduces repeated background information. RM-WAM therefore compresses recent visual history before it enters the WAM backbone.

Specifically, we select the K most recent visual frames from the memory set $\mathcal{M}_t$ as the short visual memory \(\mathcal{M}_t^s\) and encode them together with the current observation by a shared VAE into spatial patch tokens, producing $\mathbf{P}\in\mathbb{R}^{B\times(K+1)\times N\times d_v}$, where $B$ is the batch size, $N$ is the number of visual patches, and $d_v$ is the video-token dimension. Instead of flattening them into a global token sequence of length $(K+1)N$, we rearrange them as $\mathbf{S}\in\mathbb{R}^{(BN)\times(K+1)\times d_v}$, where each spatial position forms a short temporal sequence.  A shared temporal encoder aggregates recent observations at each spatial position and produces a history-conditioned representation for the current timestep. Its projected output is denoted as the short visual memory representation $\mathbf{R}_t^s$. A short-memory utility gate then updates the current visual tokens as
\begin{equation}
\begin{aligned}
\widetilde{\mathbf{X}}^{\mathrm{cur}}
&=
\mathbf{X}^{\mathrm{cur}}
+
g_t^s\mathbf{R}_t^s,
\qquad
g_t^s
&=
\sigma\left(
G_s\left(
\mathbf{X}^{\mathrm{cur}},
\mathbf{R}_t^s,
c
\right)
\right),
\end{aligned}
\end{equation}
where $\mathbf{X}^{\mathrm{cur}}$ and $\widetilde{\mathbf{X}}^{\mathrm{cur}}$ are the current visual tokens before and after memory injection, $G_s$ is the short-memory utility adapter, $\sigma(\cdot)$ is the sigmoid function, and $g_t^s\in[0,1]$ is the utility coefficient of recent visual history under the current prediction context.

\subsubsection{\textbf{Latent Event Memory}}

Short memory mainly supplements the current interaction state, while long-horizon manipulation also depends on state transitions that have already happened. To align memory with future-latent prediction in WAM, we construct event memory in the shared latent space without using image-level labels or discrete event detectors.

We sample historical visual latent states at an interval of 16 timesteps and construct $K_e$ transition-level event candidates. Each candidate $\mathbf{U}_i$ concatenates the latent states before and after the transition, denoted as $\mathbf{Z}_i^{\mathrm{before}}$ and $\mathbf{Z}_i^{\mathrm{after}}$, together with the signed change $\mathbf{Z}_i^{\mathrm{after}}-\mathbf{Z}_i^{\mathrm{before}}$ and its absolute value. This keeps both state context and transition content. After adding temporal position embeddings, the candidates are encoded into latent event representations:
\begin{equation}
  [\mathbf{e}_1,\ldots,\mathbf{e}_{K_e}]
  \in
  \mathbb{R}^{B\times K_e\times d_e},
  \end{equation}
where $B$ is the batch size, $K_e$ is the number of transition-level event candidates, $d_e$ is the event-representation dimension, and $\mathbf{e}_i$ denotes the encoded representation of the $i$-th latent transition. To obtain a current-state-conditioned latent event memory, we map the current visual to a query $\mathbf{q}_t$ and each latent event $\mathbf{e}_i$ to a key-value pair $(\mathbf{k}_i,\mathbf{v}_i)$, then softly aggregate the event values:
\begin{equation}
\begin{aligned}
r_i
&=
\operatorname{Softmax}_{i}
\left(
\frac{\mathbf{q}_t^\top\mathbf{k}_i}{\sqrt{d_r}}
\right),\\
\mathcal{M}_t^e
&=
\operatorname{Proj}_o
\left(
\sum_{i=1}^{K_e}
r_i\mathbf{v}_i
\right),
\end{aligned}
\end{equation}
where $d_r$ is the query-key dimension, $r_i$ is the normalized aggregation weight, $\operatorname{Proj}_o$ maps the aggregated value back to the WAM visual dimension, and $\mathcal{M}_t^e$ is the latent event memory conditioned on the current state. The attention weights $r_i$ reflect the relevance of each historical transition to the current state, but do not directly determine whether it should influence the current prediction.

Latent event memory is used to modulate the future visual latent tokens that model world dynamics. Specifically, we adopt an AdaLN-Zero style modulation to inject the aggregated event representation $\mathcal{M}_t^e$ into the video branch. At the $l$-th modulation layer, $\mathcal{M}_t^e$ is projected to adaptive scale and shift parameters:
\begin{equation}
\begin{aligned}
[\boldsymbol{\eta}_l,\boldsymbol{\beta}_l]
&=
W_l\mathcal{M}_t^e,\\
\mathbf{D}_l
&=
\operatorname{Adapter}_l
\left(
\operatorname{LN}(\mathbf{X}_l^{\mathrm{fut}})
\odot
(1+\boldsymbol{\eta}_l)
+
\boldsymbol{\beta}_l
\right),
\end{aligned}
\end{equation}
where $\mathbf{X}_l^{\mathrm{fut}}$ denotes the future visual latent tokens at the $l$-th modulation layer, $W_l$ is a learnable linear projection that maps the latent event memory to the modulation parameters, and $\boldsymbol{\eta}_l$ and $\boldsymbol{\beta}_l$ denote the adaptive scale and shift, respectively. $\operatorname{LN}(\cdot)$ denotes layer normalization, $\odot$ denotes element-wise multiplication, and $\operatorname{Adapter}_l$ is a lightweight projection module that produces the event-conditioned modulation term $\mathbf{D}_l$.

To control the contribution of event memory under the current prediction context, we further introduce an event-memory gate:
\begin{equation}
\begin{aligned}
g_t^e
&=
\sigma
\left(
G_e
(\mathbf{X}^{\mathrm{cur}},\mathcal{M}_t^e,c)
\right),\\
\widetilde{\mathbf{X}}_l^{\mathrm{fut}}
&=
\mathbf{X}_l^{\mathrm{fut}}
+
g_t^e\mathbf{D}_l,
\end{aligned}
\end{equation}
where $\mathbf{X}^{\mathrm{cur}}$ denotes the current visual representation, $c$ is the language instruction, $G_e$ is the event-memory gate and $\sigma(\cdot)$ denotes the sigmoid function. The resulting coefficient $g_t^e\in[0,1]$ controls the strength of event-memory injection, while $\widetilde{\mathbf{X}}_l^{\mathrm{fut}}$ denotes the future visual latent tokens after modulation. The gate is learned with counterfactual memory supervision, allowing the contribution of latent event memory to adapt to its utility under the current prediction context.

\vspace{-2mm}
\subsection{Training and Inference}
RM-WAM is trained with joint world-action modeling, future-transition supervision for latent event memory, and memory-utility supervision. The overall objective is
\begin{equation}
\mathcal{L}
=
\mathcal{L}_{\mathrm{act}}
+
\lambda_v \mathcal{L}_{\mathrm{vid}}
+
\lambda_e \mathcal{L}_{\mathrm{evt}}
+
\lambda_u \mathcal{L}_{\mathrm{utility}},
\end{equation}
where $\mathcal{L}_{\mathrm{act}}$ and $\mathcal{L}_{\mathrm{vid}}$ are the flow-matching losses for action generation and future visual dynamics, $\mathcal{L}_{\mathrm{evt}}$ encourages latent event memory to retain historical transitions related to future world evolution, and $\mathcal{L}_{\mathrm{utility}}$ provides memory-utility supervision from counterfactual memory conditions. The coefficients $\lambda_v$, $\lambda_e$, and $\lambda_u$ balance the video, event, and utility losses.

For the WAM backbone, we use flow matching to jointly model future action chunks and future visual latents. Let $\mathbf{y}$ denote either the future action chunk $\mathbf{a}_{t:t+H_a-1}$ or the future visual latent $\mathbf{z}_{t+1:t+H_v}$. During training, we sample noise $\boldsymbol{\epsilon}\sim\mathcal{N}(\mathbf{0},\mathbf{I})$ and a time step $\tau\sim\mathcal{U}(0,1)$, and define the interpolation path
\begin{equation}
\mathbf{y}_{\tau}
=
(1-\tau)\mathbf{y}
+
\tau\boldsymbol{\epsilon}.
\end{equation}

Conditioned on the latent world representation $\mathbf{s}_t$, the conditional velocity field $f_\theta$ predicts the velocity along this path:
\begin{equation}
\mathcal{L}_{\mathrm{FM}}(\mathbf{y})
=
\mathbb{E}_{\mathbf{y},\boldsymbol{\epsilon},\tau}
\left[
\left\|
f_\theta(\mathbf{y}_{\tau},\tau,\mathbf{s}_t)
-
(\boldsymbol{\epsilon}-\mathbf{y})
\right\|_2^2
\right],
\end{equation}
here $f_\theta$ is the conditional velocity predictor and $\mathcal{L}_{\mathrm{FM}}$ is the flow-matching loss for a target variable $\mathbf{y}$. The action and future-visual losses are then:
\vspace{-2mm}
\begin{equation}
\mathcal{L}_{\mathrm{act}}
=
\mathcal{L}_{\mathrm{FM}}
\left(
\mathbf{a}_{t:t+H_a-1}
\right),
\qquad
\mathcal{L}_{\mathrm{vid}}
=
\mathcal{L}_{\mathrm{FM}}
\left(
\mathbf{z}_{t+1:t+H_v}
\right),
\end{equation}
where $\mathbf{z}_{t+1:t+H_v}$ denotes the clean future visual latent sequence used as the video prediction target. Video tokens and action tokens interact in the MoT backbone, so future visual latent supervision shapes the world representation used for action generation. We also apply a causal mask consistent with the prediction targets, so future visual latents are used only as supervision for the video branch and are not exposed as visible inputs to the action prediction branch.

To encourage latent event memory to retain transitions that are predictive of future world evolution, we introduce a future-transition objective:
\begin{equation}
\widehat{\mathbf{y}}_{\mathrm{evt}}
=
h_{\mathrm{evt}}(\mathbf{M}_t^e),
\qquad
\mathcal{L}_{\mathrm{evt}}
=
\operatorname{SmoothL1}
\left(
\widehat{\mathbf{y}}_{\mathrm{evt}},
\operatorname{sg}(\mathbf{y}_{\mathrm{evt}})
\right),
\end{equation}
where $\mathbf{y}_{\mathrm{evt}}$ is a latent-transition target constructed from the current and future visual latent sequences, $\widehat{\mathbf{y}}_{\mathrm{evt}}$ is its prediction from the event head $h_{\mathrm{evt}}$, and $\operatorname{sg}(\cdot)$ denotes stop-gradient. Future visual latents are used only to define this training target and do not participate in building the event memory itself.

Each counterfactual memory condition provides a sample-level soft utility target $q_v\in[0,1]$. RM-WAM uses this target to supervise the utility adapters of both short-term visual state memory and latent event memory:
\begin{equation}
\mathcal{L}_{\mathrm{utility}}
=
\operatorname{BCE}(g_t^s,q_v)
+
\operatorname{BCE}(g_t^e,q_v),
\end{equation}
where $\operatorname{BCE}$ denotes binary cross-entropy. In the current implementation, $g_t^s$ and $g_t^e$ share the same sample-level soft utility target derived from the selected counterfactual memory condition. We use a shared target because both memory branches are constructed under the same counterfactual history condition, while their branch-specific gating networks independently determine the injection strength from their respective memory representations.

During inference, RM-WAM maintains a causal visual history buffer for each episode. After the robot executes an action chunk and receives a new observation, the observation is appended to the buffer in temporal order, and memory is sampled with the same temporal offsets as in training at the next decision step. When valid history is insufficient at the beginning of an episode, missing positions are masked instead of being filled by copying the current frame. Since causal attention masks already prevent the action branch from accessing future visual information during training, future video generation can be omitted during inference without affecting action prediction.

\section{Experiments}
\subsection{Experimental Setup}
\subsubsection{\textbf{Model architecture}} RM-WAM adopts a dual-branch architecture for joint modeling of future visual and robot actions. The video branch uses Wan2.2-TI2V-5B \cite{wan2025wan} as the backbone for world prediction, while the action branch employs a 30-layer Diffusion Transformer (DiT) \cite{peebles2023scalable} to generate continuous action chunks. The two branches interact through a Mixture-of-Transformers (MoT). In addition, a one-layer temporal Transformer encodes recent visual latent tokens to construct short-term visual memory, while another one-layer Transformer encodes historical latent transitions to form latent event memory. For the two memory types, lightweight two-layer MLP gating modules are employed to control memory injection, with their strengths learned through counterfactual memory supervision.

\subsubsection{\textbf{Benchmarks}} We evaluate RM-WAM on two simulated benchmarks, LIBERO \cite{liu2023libero} and RMBench \cite{chen2026rmbench}. LIBERO evaluates language-conditioned manipulation across four suites: LIBERO-Spatial, LIBERO-Object, LIBERO-Goal, and LIBERO-Long, covering spatial, object-centric, goal-conditioned, and long-horizon tasks. RMBench focuses on memory-dependent manipulation under partial observability, where reliable execution requires the model to retain and utilize historical visual states and interaction events over extended temporal horizons. Together, the two benchmarks assess both the general manipulation capability of RM-WAM and its effectiveness in long-horizon tasks.

\subsubsection{\textbf{Baselines}} We compare RM-WAM with four representative methods. Diffusion Policy (DP)\cite{chi2025diffusion} is included as an action-generation baseline without world modeling. $\pi_{0.5}$ \cite{intelligence2025pi_} represents a strong vision-language-action policy for continuous robot control. FastWAM \cite{yuan2026fast} serves as a world--action modeling baseline that jointly learns visual and robot actions. HiMem-WAM \cite{sun2026himem} is the most closely related memory-augmented world-action model and introduces hierarchical memory gating for long-horizon manipulation. All methods are evaluated using the same data and evaluation protocol.

\subsubsection{\textbf{Implementation Details}} 
Short-term memory encodes the four most recent frames, while event memory samples visual latents every 16 timesteps to construct four transition-level candidates. Counterfactual supervision is implemented by modifying the historical sampling pools of short and event memory while keeping their memory lengths fixed. The reference, stale, stage, and cross-episode conditions are assigned soft utility targets of $1.0$, $0.01$, $0.01$, and $0.01$, respectively. These values are used as soft supervision priors to distinguish causally aligned history from deliberately perturbed memory conditions, and counterfactual memory samples account for 50\% of the training data.

Multi-view observations are concatenated into a single image before being fed into the model. Robot actions are represented in 7-DoF, with an action horizon of 32. Models are trained for 10 epochs with a batch size of 16 and gradient accumulation of 1, using AdamW with an initial learning rate of $1\times10^{-4}$ and a weight decay of $1\times10^{-2}$. During training, gradients from the action branch to the video branch are stopped. At inference, RM-WAM maintains a causal history buffer that is updated with incoming observations, and unavailable historical states at the beginning of an episode are masked out. For action generation, we use classifier-free guidance (CFG)~\cite{2022arXiv220712598H} with a guidance scale of 1.0. 






\subsection{Simulation}
\subsubsection{\textbf{Task}} For both LIBERO and RMBench, each task uses 50 expert demonstrations for training, and all models are trained for 10 epochs under the same configuration. On LIBERO, we evaluate all 40 tasks across the four suites, performing 50 evaluation rollouts per task with different environment initialization seeds, and report the average success rate. On RMBench, we evaluate the nine bimanual manipulation tasks with 100 evaluation rollouts per task under different environment initialization seeds and also report the average success rate.

\begin{table}[H]
\centering
\caption{\textbf{Comparisons with other baselines on Libero benchmark.} Success rates(\%) are reported, and bold denotes the best performance among all methods.}
\vspace{-3mm}
\label{tab:category_results}
\setlength{\tabcolsep}{6pt}
\renewcommand{\arraystretch}{1.15}
\resizebox{1\linewidth}{!}{
\begin{tabular}{lccccc}
\toprule
\textbf{Method} & \textbf{Spatial} & \textbf{Object} & \textbf{Goal} & \textbf{Long} & \textbf{Avg.} \\
\midrule
DP             & 78.3\% & 92.5\% & 68.3\% & 50.5\% & 72.4\% \\
$\pi_{0.5}$    & 98.8\% & 98.2\% & 98.0\% & 92.4\% & 96.9\% \\
HiMem-WAM      & 98.2\% & 99.8\% & 98.4\% & 94.5\% & 97.7\% \\
FastWAM       & 98.2\% & 100.0\% & 97.0\% & 95.2\% & 97.6\% \\
\textbf{RM-WAM (Ours)}
               & \textbf{98.8\%} & \textbf{100.0\%} & \textbf{98.2\%} & \textbf{97.0\%} & \textbf{98.5\%} \\
\bottomrule
\end{tabular}}
\end{table}

\subsubsection{\textbf{Evaluation Results}} As shown in Tables~\ref{tab:category_results} and~\ref{tab:main_results}, RM-WAM achieves the best overall performance on both LIBERO and RMBench. On LIBERO, RM-WAM reaches an average success rate of 98.5\%, with the most notable improvement on LIBERO-Long, indicating that short-term interaction cues and historical state transitions provide richer contextual support for long-horizon decision making than relying solely on the current observation. On RMBench, RM-WAM achieves an average success rate of 55.8\% and substantially outperforms HiMem-WAM. This indicates that for strongly history-dependent tasks, the key is to identify which past states and interaction events are useful for the current decision. By combining dual-scale memory with counterfactual supervision for selective memory utilization, RM-WAM achieves more gains on RMBench.


\begin{table}[t]
\centering
\caption{\textbf{Comparisons with baselines on RMBench.} Success rates (\%) are reported, and bold denotes the best performance.}
\vspace{-3mm}
\label{tab:main_results}
\setlength{\tabcolsep}{3.5pt}
\renewcommand{\arraystretch}{1.08}
\resizebox{\columnwidth}{!}{
\begin{tabular}{lccccc}
\toprule
\textbf{Task}
& \textbf{DP}
& $\boldsymbol{\pi_{0.5}}$
& \textbf{FastWAM}
& \textbf{HiMem-WAM}
& \textbf{RM-WAM} \\
\midrule

Observe Pick Up
& 1\% & 9\% & 0\% & \textbf{28\%} & 27\% \\

Rearrange Blocks
& 0\% & 13\% & 0\% & 33\% & \textbf{65\%} \\

Put Back Block
& 0\% & 11\% & 0\% & 32\% & \textbf{70\%} \\

Swap Blocks
& 11\% & 24\% & 0\% & 38\% & \textbf{62\%} \\

Swap T
& 2\% & 15\% & 7\% & 27\% & \textbf{58\%} \\

Battery Try
& 10\% & 16\% & 20\% & 28\% & \textbf{30\%} \\

Blocks Ranking
& 10\% & 6\% & 26\% & 24\% & \textbf{75\%} \\

Cover Blocks
& 0\% & 0\% & 0\% & 19\% & \textbf{64\%} \\

Press Button
& 0\% & 0\% & 0\% & 8\% & \textbf{51\%} \\

\midrule
\textbf{Avg.}
& 3.8\% & 10.4\% & 5.9\% & 26.3\% & \textbf{55.8\%} \\

\bottomrule
\end{tabular}}
\end{table}

\subsection{Real World}

\subsubsection{\textbf{Task}}
\begin{figure}[H]
    \centering
    \includegraphics[width=0.45\textwidth]{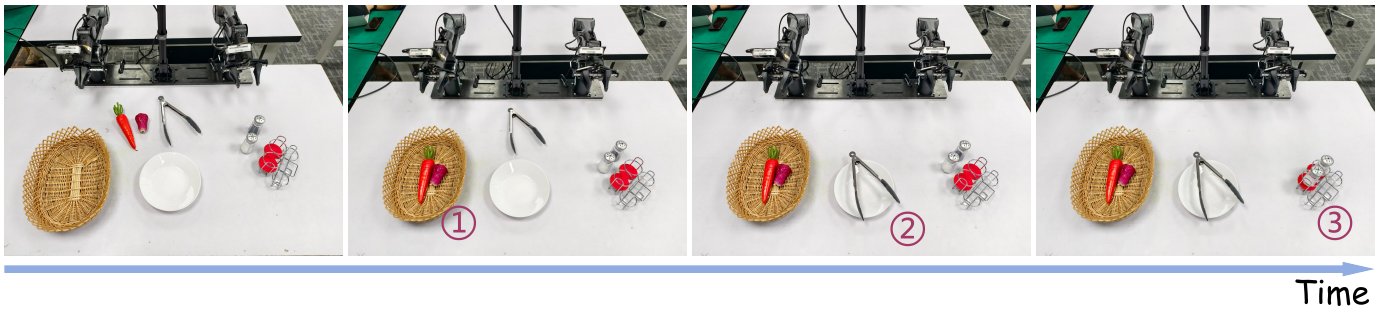}
    \vspace{-4mm}
    \caption{\textbf{Real-world long-horizon tabletop organization task.} The task consists of three sequential manipulation subtasks, denoted by \textcircled{1}--\textcircled{3}, requiring the robot to complete multiple object organization steps over an extended temporal horizon.}
    \label{fig:real-world}
\end{figure}

We further evaluate RM-WAM on a real-world dual-arm robotic platform consisting of two AGILEX PIPER manipulators, an Intel RealSense D435i
RGB-D camera. We design a long-horizon \textit{Organize Table} task that requires the robot to complete a sequence of object manipulation steps within a single episode. As shown in Fig. \ref{fig:real-world}, the task consists of multiple temporally dependent stages and therefore requires the policy to maintain and utilize historical information throughout execution. We collect 500 expert demonstrations for training, each covering the complete task trajectory from the initial scene to the final organized state. During evaluation, the initial object configurations are randomized, and the robot is required to complete the entire multi-stage task. For real-world inference, the policy predicts an action chunk of 32 and replans every 8 control steps at a control frequency of 30 Hz. To further evaluate robustness and generalization in real-world settings, we consider three testing settings: \textit{Standard} evaluates the complete task under normal initial configurations. \textit{Disturbance} introduces external interventions during execution, such as manually removing an object that has already been placed at its target location. \textit{Unseen} introduces objects that are absent from the training data. We conduct 30 rollouts for each setting and report the overall average success rate.

\subsubsection{\textbf{Evaluation Results}} We compare RM-WAM with FastWAM, which follows a similar world-action modeling paradigm, to examine the effectiveness of the proposed reliable memory mechanism in physical environments. As shown in Table~\ref{tab:real_world_results}, RM-WAM consistently outperforms FastWAM across all three settings. Under \textit{Standard}, RM-WAM achieves a success rate of 90.0\%, improving over FastWAM by 10 percentage points, indicating more stable execution in long-horizon multi-stage tasks. Under \textit{Disturbance}, RM-WAM reaches 83.3\%, compared with 70.0\% for FastWAM. This suggests that when external interventions alter task progress, relying mainly on the current observation can lead to incorrect stage estimation, whereas RM-WAM can leverage historical interactions to better reassess the current task state and recover from disturbances. Under \textit{Unseen}, RM-WAM achieves 86.7\%, compared with 76.7\% for FastWAM, showing that reliable historical context remains beneficial when generalizing to unseen objects. 
\vspace{-2mm}
\begin{table}[H]
\centering
\caption{\textbf{Real-world evaluation results.} Success rates (\%) are reported, and bold denotes the best performance.}
\vspace{-3mm}
\label{tab:real_world_results}
\setlength{\tabcolsep}{7pt}
\renewcommand{\arraystretch}{1.1}
\begin{tabular}{lccc}
\toprule
\textbf{Method}
& \textbf{Standard}
& \textbf{Disturbance}
& \textbf{Unseen} \\
\midrule
FastWAM
& 80.0
& 70.0
& 76.7 \\
\textbf{RM-WAM}
& \textbf{90.0}
& \textbf{83.3}
& \textbf{86.7} \\
\bottomrule
\end{tabular}
\end{table}

\vspace{-3mm}
\subsection{Memory Analysis and Visualization}
\subsubsection{\textbf{Short-Term Visual Memory}} 
\begin{figure}[H]
    \centering
    \includegraphics[width=0.45\textwidth]{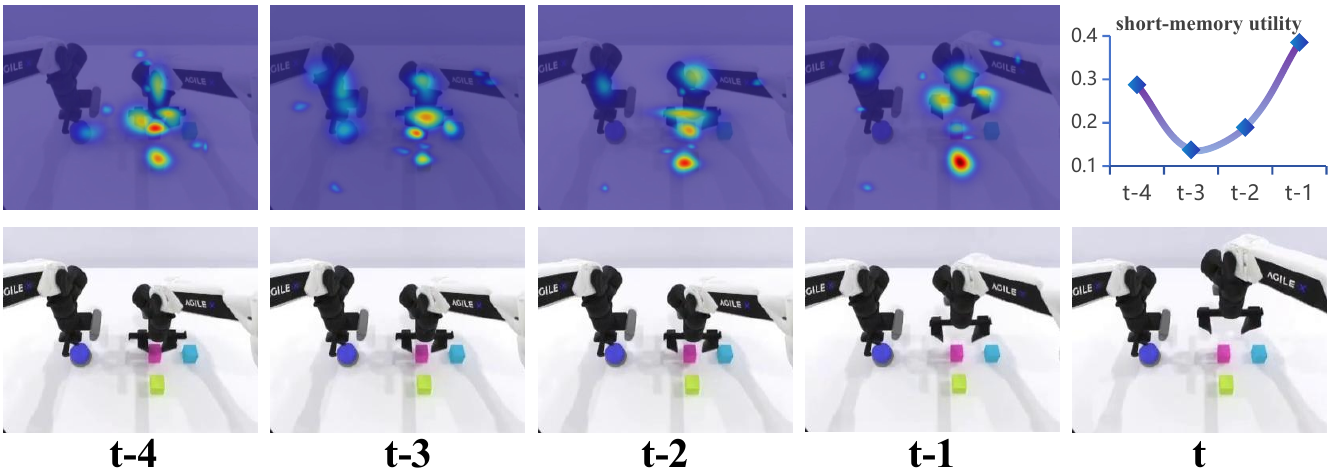}
    \vspace{-3mm}
    \caption{
    \textbf{Visualization of short-term visual memory in RM-WAM.} The top row shows spatial attention over recent historical observations, and the bottom row shows the corresponding raw observations together with the current frame. The right panel shows the short-memory utility coefficient learned through counterfactual memory supervision.
    }
    \label{fig:short_memory_vis}
\end{figure}

To understand the effectiveness of the short-term memory in RM-WAM, we visualize its spatial attention and the corresponding utility coefficient on the \textit{Blocks Ranking Try} task in RMBench. As shown in Fig.~\ref{fig:short_memory_vis}, at the current timestep $t$, the robot has just completed placing the pink block and should subsequently interact with the green block. The short-term memory mainly attends to manipulation-relevant regions, including the gripper, the previously manipulated pink block, and the green block corresponding to the next interaction, providing historical cues for identifying the subsequent manipulation target. We further visualize the short-memory utility coefficient learned through counterfactual memory supervision. As the task transitions from placing the pink block to manipulating the green block, the utility increases toward the current decision and reaches its highest value at $t-1$. This indicates that RM-WAM does not uniformly fuse recent observations, but instead adaptively evaluates the utility of short-term history according to the current task state.

\vspace{-3mm}
\vspace{-2mm}

\subsubsection{\textbf{Latent Event Memory}} 
\begin{figure}[H]
    \centering
    \includegraphics[width=0.38\textwidth]{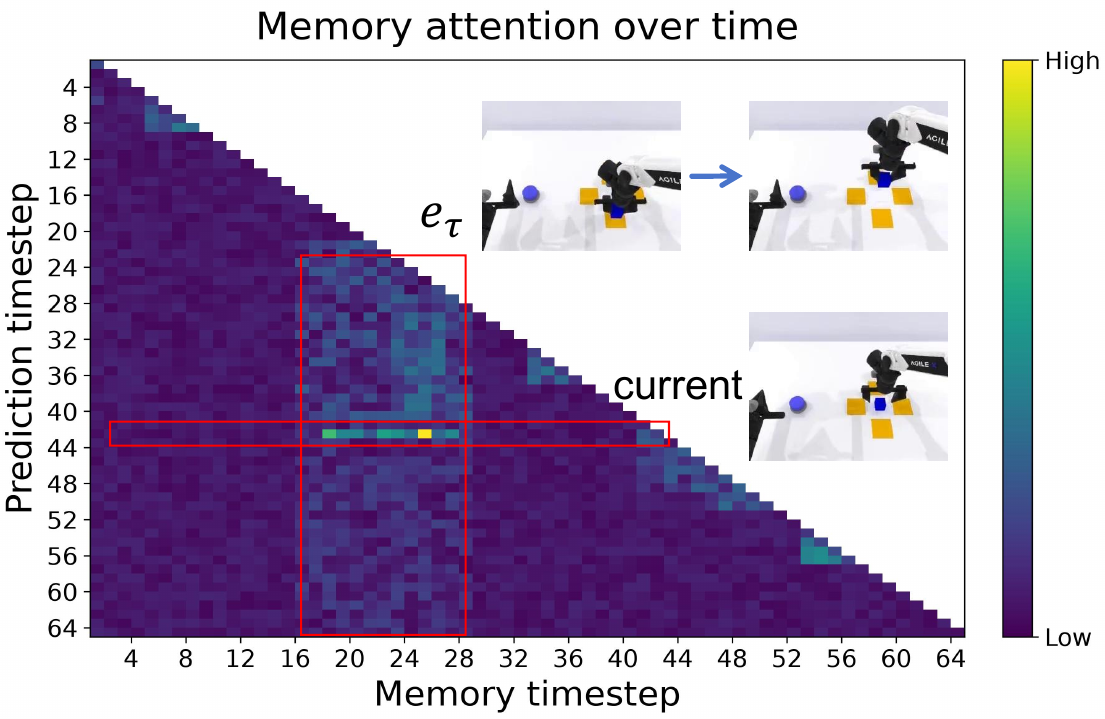}
   \caption{\textbf{Visualization of latent event memory in RM-WAM.} The model attends to earlier event $e_{\tau}$ when returning the block to its original position.}
    \label{fig:event-attention}
\end{figure}

To analyze how RM-WAM utilizes latent event memory during long-horizon manipulation, we visualize the memory attention over time on the \textit{Put Back Block} task in RMBench. As shown in Fig.~\ref{fig:event-attention}, when the robot needs to return the block to its original position, the model assigns noticeably higher attention to an earlier historical interval corresponding to the event in which the block was initially picked up from that location. The result provides qualitative evidence that latent event memory captures meaningful interaction events and supports history-dependent decision making in long-horizon tasks.
\subsubsection{\textbf{Memory-Conditioned Future Prediction}} 
\begin{figure}[H]
    \centering
    \includegraphics[width=0.4\textwidth]{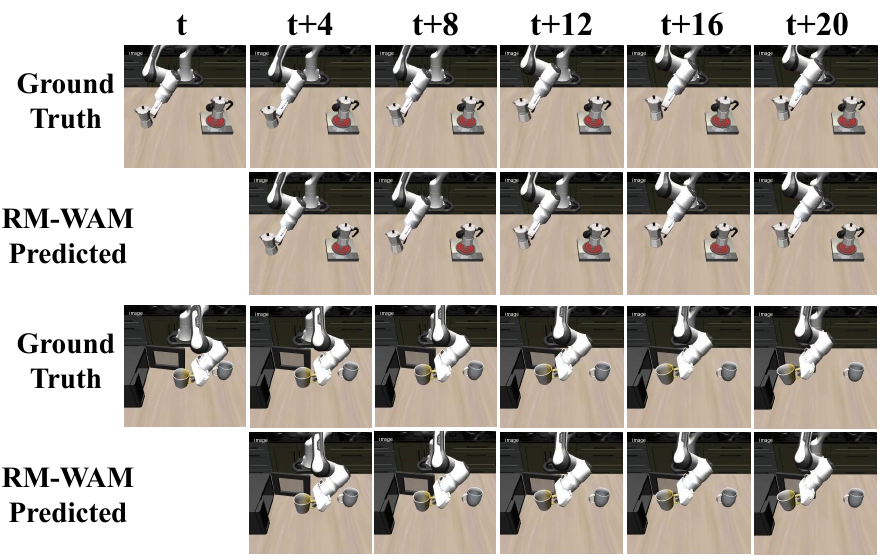}
    \caption{\textbf{Visualization of future evolution on LIBERO-Long.}}
    \label{fig:future_prediction}
\end{figure}

To further investigate how reliable memory contributes to world modeling, we visualize the future state evolution of RM-WAM on two tasks from LIBERO-Long. Given the current and historical observations, RM-WAM predicts future visual states over a long horizon. As shown in Fig.~\ref{fig:future_prediction}, the predicted trajectories closely match the ground-truth observations across different future timesteps, preserving the evolution of object configurations and robot-object interactions. These results indicate that RM-WAM can effectively generate temporally consistent and task-relevant future dynamics.

\subsection{Ablation Studies}
We conduct an ablation study on LIBERO to analyze the short-term memory and latent event memory across tasks. As shown in Table~\ref{tab:ablation_libero}, \textit{w/o short-mem.} and \textit{w/o event-mem.} denote removing the short-term   memory and latent event memory, respectively. Removing short-term memory decreases the success rate on LIBERO-Spatial from 98.8\% to 97.8\%, indicating that recent visual history is important for preserving local spatial information. On LIBERO-Goal and LIBERO-Long, removing event memory leads to larger performance drops, with success rates decreasing to 96.8\% and 95.0\%, respectively. This is consistent with the stronger dependence of these tasks on task progress and longer-range state transitions, for which latent event memory provides more persistent historical context. Overall, RM-WAM achieves the highest average success rate, validating the complementary roles of the two memory components.

\vspace{2mm}
\begin{table}[H]
\centering
\caption{\textbf{Ablation study on LIBERO.} Success rates (\%) are reported, and bold denotes the best performance.}
\label{tab:ablation_libero}
\setlength{\tabcolsep}{4.5pt}
\renewcommand{\arraystretch}{1.1}
\resizebox{\columnwidth}{!}{
\begin{tabular}{lccccc}
\toprule
\textbf{Method}
& \textbf{Goal}
& \textbf{Object}
& \textbf{Spatial}
& \textbf{Long}
& \textbf{Avg.} \\
\midrule
w/o short-mem.
& 97.0
& 100.0
& 97.8
& 96.0
& 97.7 \\

w/o event-mem.
& 96.8
& 100.0
& 98.2
& 95.0
& 97.5 \\

\textbf{RM-WAM}
& \textbf{98.2}
& \textbf{100.0}
& \textbf{98.8}
& \textbf{97.0}
& \textbf{98.5} \\
\bottomrule
\end{tabular}}
\end{table}

\section{Conclusion}
In this paper, we propose RM-WAM, a reliable memory-augmented World Action Model for long-horizon robot manipulation. Different from existing memory augmentation methods that treat historical information as universally useful context, RM-WAM considers memory as a context-dependent predictive resource and models history through complementary short-term visual memory and latent event memory to capture local interactions and long-term state transitions. Furthermore, counterfactual memory learning is introduced to learn a context-dependent utility estimate for historical information under the current task state, enabling adaptive memory utilization for world prediction and action generation. Experiments demonstrate that RM-WAM effectively exploits reliable historical information and improves long-horizon task performance through more accurate world evolution modeling and closed-loop decision making.

\bibliographystyle{IEEEtran}
\bibliography{references}

\end{document}